\documentclass[runningheads]{llncs}
\usepackage[T1]{fontenc}
\usepackage{graphicx}
\usepackage{cite}
\usepackage{amsmath}
\usepackage{amssymb,amsfonts}
\usepackage{algorithmic}
\usepackage{textcomp}
\usepackage{xcolor}
\usepackage{booktabs} 
\usepackage{multirow}
\usepackage{array}
\usepackage{url}

\begin{document}
\title{NeuroGAN-3D: Enhancing Intrinsic Functional Brain Networks via High-Fidelity 3D Generative Super-Resolution}
\titlerunning{NeuroGAN-3D}
%
\author{
M.~Moein Esfahani\inst{1}\orcidID{0000-0003-0803-1963} \and
Sepehr Salem Ghahfarokhi\inst{2}\orcidID{0000-0002-6308-1202} \and
Mohammed Alser\inst{2}\orcidID{0000-0002-6117-3701} \and
Jingyu Liu\inst{1}\orcidID{0000-0002-1724-7523} \and
Vince Calhoun\inst{1}\orcidID{0000-0001-9058-0747}
}

\authorrunning{M. Esfahani et al.}

\institute{
\inst{2} Georgia State University, Atlanta, GA, USA\\
\email{\{ssalemghahfarokhi1,malser\}@gsu.edu}
\and
\inst{1} Tri-Institutional Center for Translational Research in Neuroimaging and Data Science (TReNDS),
Georgia State University, Georgia Institute of Technology, Emory University, Atlanta, GA, USA\\
\email{\{mesfahani1,jliu75,vcalhoun\}@gsu.edu}
}

\maketitle              
\begin{abstract}
Recent advances in neuroimaging have deepened
our understanding of the brain’s complex functional and structural organization. Among these, functional Magnetic Resonance
Imaging (fMRI) – particularly resting-state fMRI (rs-fMRI) –
has emerged as a tool for identifying biomarkers of
intrinsic brain connectivity and delineating large-scale neural
networks. These networks are typically represented as volumetric
spatial maps that capture functionally coherent brain regions
and reflect individual differences in brain activity and structure.
The spatial resolution of these maps plays important role, as it
determines the ability to localize functional units with precision,
perform reliable brain parcellation, and detect subtle, spatially
specific neurobiological alterations associated with development,
aging, or disease. Therefore, improving the effective resolution
of neuroimaging-derived maps holds significant promise for
enabling more detailed insights into brain architecture and its
relationship to behavior and pathology. To address this need, we
propose NeuroGAN-3D, a novel 3D generative super-resolution
model tailored to the computational demands of volumetric
neuroimaging. Our model leverages a generative adversarial
network architecture to enhance the spatial resolution of rs-fMRI
spatial maps, significantly outperforming a conventional baseline.

\end{abstract}
\section{Introduction}

Recent advances in neuroimaging have deepened our understanding of the brain’s complex functional and structural organization. Among these, functional Magnetic Resonance Imaging (fMRI) – particularly resting-state fMRI (rs-fMRI) – has emerged as a critical tool for identifying biomarkers of intrinsic brain connectivity and delineating large-scale neural networks \cite{zening1,p_rs_fmri}. These networks are typically represented as volumetric spatial maps where the spatial resolution is important for localizing functional units with precision and detecting subtle, spatially specific neurobiological alterations \cite{zening2,p_fmri_data}. Therefore, improving the effective resolution of these maps holds significant promise for enabling more detailed insights into brain architecture and its relationship to behavior and pathology \cite{p_sr_smalld}. However, direct high-resolution (HR) acquisition remains constrained by technical, practical, and biophysical limitations, including extended scan times and reduced signal-to-noise ratio (SNR), which has led to a growing interest in computational super-resolution (SR) techniques \cite{p_sr_gan,p_medvae}.

Traditional SR methods, like trilinear or Lanczos resampling, are computationally efficient but fundamentally limited, often producing blurred reconstructions that lack anatomical fidelity \cite{p_medvae,p_rs_fmri}. Early machine learning (ML) efforts offered modest improvements but were limited in modeling the highly nonlinear mappings between low-resolution (LR) and HR content \cite{p_medvae,p_cnn_mdl}. The emergence of deep learning, particularly convolutional neural networks (CNNs), significantly advanced the field. Early models such as SRCNN demonstrated notable gains over classical methods by learning end-to-end mappings between LR and HR domains \cite{p_cnn_sr}. More recently, Generative Adversarial Networks (GANs) have set new benchmarks in perceptual quality, with architectures like the Enhanced Super-Resolution Generative Adversarial Network (ESRGAN) achieving state-of-the-art realism in 2D natural image SR \cite{p_esrgan, p_cnn_sr}. These improvements are largely due to adversarial training and perceptual loss functions that guide reconstruction toward images that are not only pixel-accurate but also visually and structurally consistent.

Inspired by these advances, GAN-based SR methods have been increasingly explored in medical imaging applications to enhance anatomical resolution \cite{p_sr_smalld, p_sr_gan}. However, extending these approaches to volumetric neuroimaging data—especially functional brain networks derived from fMRI—introduces unique challenges. Most existing SR methods operate on a slice-by-slice 2D basis, which fails to account for three-dimensional continuity and often leads to through-plane inconsistencies. While several 3D SR models have been proposed, many are not specifically tailored for reconstructing derived functional spatial maps and do not fully exploit the architectural strengths of advanced GANs like ESRGAN \cite{p_3d_sr,p_esrgan}. Ensuring anatomical plausibility and avoiding the introduction of clinically misleading artifacts is paramount, making the design of high-fidelity SR models particularly challenging.

\begin{figure*}[ht]
  \includegraphics[width=\textwidth]{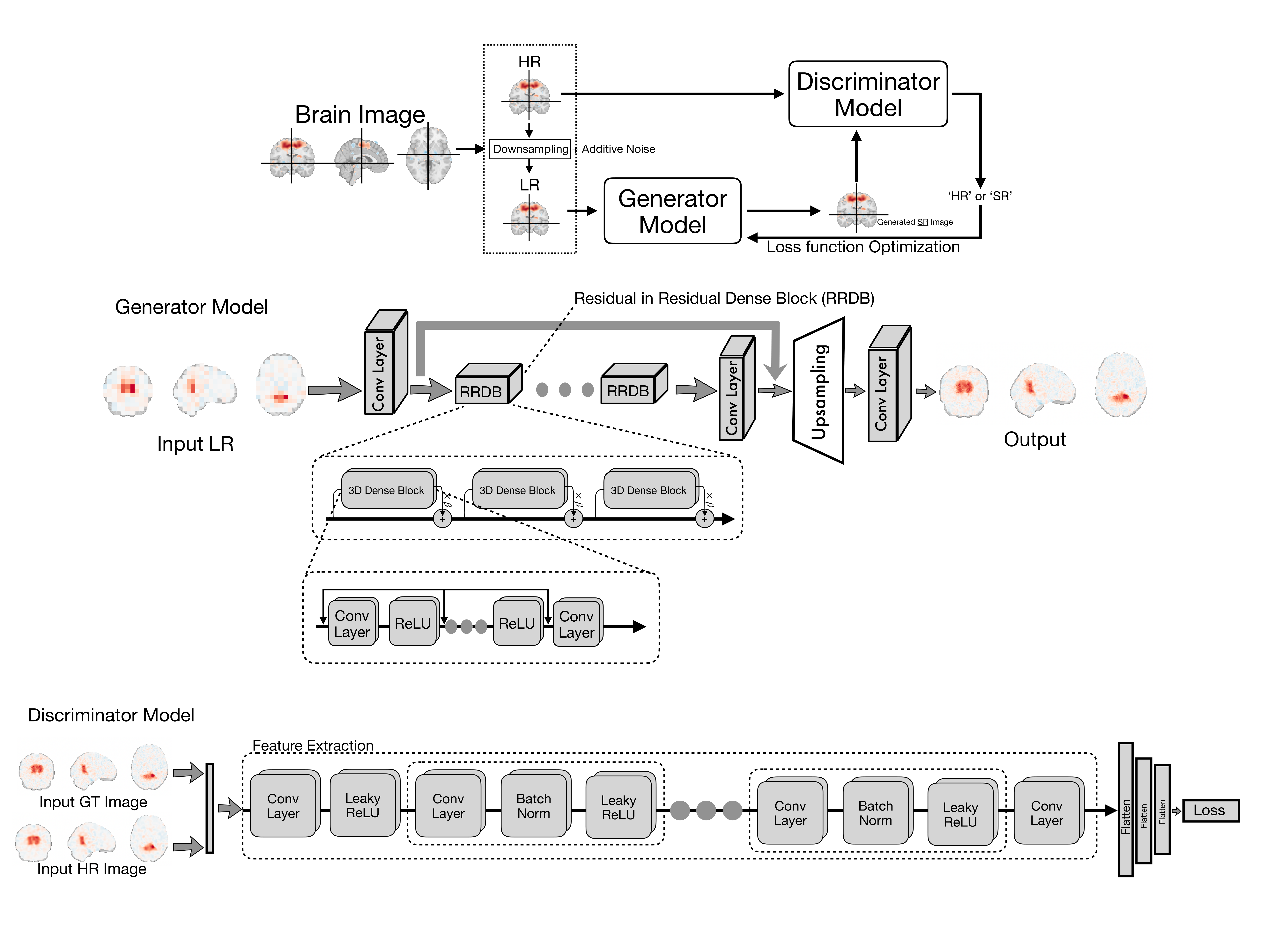}
  \caption{Overview of the proposed NeuroGAN-3D framework. The model consists of a Generator and a Discriminator. The Generator takes a low-resolution (LR) brain volume as input and uses a series of 3D Residual-in-Residual Dense Blocks (RRDBs) to reconstruct a high-resolution (HR) output. The Discriminator is trained to differentiate between real HR volumes (Ground Truth) and the generated HR volumes, providing adversarial feedback to optimize the Generator.}
\label{fig:proposed_method}
\end{figure*}

To address these challenges, we propose NeuroGAN-3D, a novel framework specifically designed for the high-fidelity super-resolution of 3D neuroimaging data, with a focus on resting-state fMRI spatial maps. An overview of our proposed model is presented in Fig. \ref{fig:proposed_method}. The main contributions of this work are as follows:
\begin{itemize}
    \item We propose a novel 3D Generative Adversarial Network (GAN) architecture, NeuroGAN-3D, specifically tailored to the volumetric nature and computational demands of neuroimaging data. The generator features 3D Residual-in-Residual Dense Blocks (RRDB-3D) to effectively learn complex spatial features.
    \item We introduce a custom composite loss function that combines a pixel-wise fidelity term with a perceptually-driven, SSIM-based loss. This approach ensures anatomical plausibility and mitigates common GAN artifacts without relying on pre-trained networks from irrelevant domains.
    \item We demonstrate the successful application of NeuroGAN-3D for enhancing the resolution of rs-fMRI spatial maps, showing that our model significantly improves both quantitative accuracy and qualitative realism compared to a conventional interpolation baseline.
\end{itemize}

\section{Methods}
The proposed NeuroGAN-3D framework is designed to achieve high-fidelity super-resolution for volumetric neuroimaging data. Built upon a Generative Adversarial Network (GAN) architecture specifically adapted for 3D volumes, the model is optimized to preserve inter-slice continuity and accurately reconstruct complex three-dimensional anatomical structures. Our design is primarily inspired by the Enhanced Super-Resolution Generative Adversarial Network (ESRGAN) \cite{p_esrgan}, with key adaptations tailored to the unique challenges of neuroimaging data.
NeuroGAN-3D comprises two core components: a 3D generator network $(G)$, which performs upsampling of low-resolution $(G)$ volumetric input to produce super-resolved $(SR)$ outputs; and a 3D discriminator network $(D)$, which is trained to distinguish between the generated SR volumes and the corresponding ground-truth HR volumes. Both the generator and discriminator architectures, along with the associated loss functions, are specifically modified to support volumetric data and to enable accurate, artifact-free reconstruction of anatomical and functional features in brain imaging.

\subsection{Generator Architecture $G$}
The generator in Model is designed to process and learn from 3D volumetric data, aiming to generate and reconstruct fine anatomical details. Its architecture is based on an enhanced version of the Residual-in-Residual Dense Block (RRDB) structure, adapted for 3D inputs, which we term RRDB-3D.
The model designed to transform a low-resolution (LR) 3D neuroimage volume $V_{LR} \in \mathbb{R}^{D_{LR} \times H_{LR} \times W_{LR}}$ into High-resolution (HR) volume $V_{SR} \in \mathbb{R}^{ D_{HR} \times H_{HR} \times W_{HR}}$. where D,H,W represent depth, height, and width respectively. in this study we assumed the model has input of 1 channel.

The architecture begins with an initial 3D convolutional layer (Conv3D) to extract low-level features $F_0$ from $V_{LR}$. The core of the generator comprises $N_{RRDB}$ as
3D Residual-in-Residual Dense Blocks (RRDB-3D). Each RRDB-3D block contains several 3D Dense Blocks (DB-3D). Within each DB-3D, multiple Conv3D layers (kernel 3×3×3, LeakyReLU activation) are densely connected, where feature maps from all preceding layers are concatenated and passed to the subsequent layer. The output of each DB-3D is scaled by a factor $\beta ( 0.2)$ and added back to its input, forming a residual connection. This structure is repeated within each RRDB-3D block. The output of the stacked RRDB-3D modules $F_{deep}$ is added to the initial features $F_0$ via a global residual connection as $F_{body} = F_0 + F_{deep}$. next step, a 3D upsampling module then increases the spatial resolution of $F{body}$. This is achieved using either 3D transposed convolutions or a sequence of trilinear interpolation followed by Conv3D layers to reach the target HR dimensions. Finally, one or more Conv3D layers refine the upsampled features, with a concluding Conv3D layer reconstructing the HR volume as $V_{HR}$.

\subsection{Discriminator Architecture $D$}
The discriminator in the model is designed as a 3D PatchGAN. Instead of discriminating and classifying an entire volume as real or fake, it classifies overlapping 3D patches of the input volume, providing more localized feedback to the generator and encouraging finer detail generation. The discriminator is designed to differentiate between real HR volumes and generated SR volumes by assessing local 3D patches. It takes either $V_{HR}$ or $V_{SR}$ as input.
The architecture consists of a series of Conv3D layers with increasing filter counts and decreasing spatial resolution, achieved through strided convolutions (stride $3 \times 3 \times 3$). Except for the input and output layers, these convolutions are typically followed by spectral normalization applied to Conv3D weights and LeakyReLU activation functions. The final \texttt{Conv3D} layer outputs a 3D score map, where each scalar value corresponds to the "realness" classification of an overlapping 3D patch in the input volume.
We also employ a Relativistic average Discriminator (RaD) strategy. The discriminator is trained to assess whether a real image is relatively more realistic than a fake one, and vice versa, rather than predicting an absolute realness score. If $C(x)$ denotes the raw discriminator output (pre-sigmoid) for an input $x$, the RaD output for a real image $x_r$ relative to fake images $x_f$ is $\sigma(C(x_r) - \mathbb{E}_{x_f}[C(x_f)])$, and for a fake image $x_f$ relative to real images $x_r$ is $\sigma(C(x_f) - \mathbb{E}_{x_r}[C(x_r)])$, where $\sigma$ is the sigmoid function.

\subsection{Loss Functions}
The NeuroGAN-3D generator $G$ is trained using modified loss function $L_G$, which is a weighted sum of pixel-wise, perceptual, and adversarial losses:
\begin{equation}
L_G = \lambda_{pixel} L_{pixel} + \lambda_{perc} L_{perc} + \lambda_{adv} L_{adv}^G
\end{equation}
where $\lambda_{pixel} = 1$, $\lambda_{perc} = 0.8$, and $\lambda_{adv} = 0.8$ are the respective weighting coefficients. These weights were determined empirically through a process of trial and error on the validation set. Our rationale was to anchor the model with a strong pixel-wise loss ($\lambda_{pixel} = 1$) to ensure fundamental content fidelity, while assigning slightly lower but significant weights to the perceptual and adversarial terms. This balance encourages the generator to produce visually plausible details without straying too far from the ground truth, thereby preventing the introduction of major artifacts.

\subsubsection{Pixel-wise Loss ($L_{pixel}$)}
An L1 loss (Mean Absolute Error) is used to ensure content fidelity between the generated SR volume $G(V_{LR})$ and the ground-truth HR volume $V_{HR}$:
\begin{equation}
L_{pixel} = \mathbb{E}_{V_{LR}} [ \lVert G(V_{LR}) - V_{HR} \rVert_1 ]
\end{equation}

\subsubsection{Perceptual Loss ($L_{perc}$)}
To enhance the virtual fidelity of SR outputs without relying on pre-trained deep networks, we introduce modified perceptual loss function.
This function is directly derived from the function of the Structural Similarity Index Measure (SSIM). It introduced to better capture perceptually significant features in data.
Given the 3D volumetric data, this loss is computed by comparing 2D slices like axial, sagittal or coronal extracted from volume $G(V_{\text{LR}})$ and HR volume $V_{\text{HR}}$.
The perceptual loss, $L_{\text{perc}}$, is then used to minimize the dissimilarity based on SSIM:
\begin{equation}
L_{\text{perc}} = \sum_{\text{plane}} \mathbb{E}_{\text{slices}} [ 1 - \text{SSIM}(\text{Slice}_{\text{SR}}, \text{Slice}_{\text{HR}}) ]
\label{eq:perceptual_loss_ssim}
\end{equation}
this loss function aims to specific adaptations to SSIM. It mainly focuses on fidelity and multi-scale analysis without need to train another model for evaluation as it used in other published loss functions\cite{p_3desrgan, p_esrgan}. This allows for perceptually aware and fidelity optimization by directly evaluating the structural similarity between the generated and target images, bypassing the need for features from external pre-trained models.

\subsubsection{Adversarial Loss ($L_{adv}$)}
The adversarial loss encourages the generator to produce SR volumes indistinguishable from real HR volumes. Using the Relativistic average Discriminator (RaD) formulation, the adversarial loss for the generator ($L_{adv}^G$) is:

\begin{equation}
\begin{split}
L_{adv}^G = -\mathbb{E}_{x_r}[\log(1 - \sigma(C(x_f) - \mathbb{E}_{x_r}[C(x_r)]))] \\ - \mathbb{E}_{x_f}[\log(\sigma(C(x_r) - \mathbb{E}_{x_f}[C(x_f)]))]
\end{split}
\label{eq:adv_loss}
\end{equation}

where $x_r$ denotes real HR volumes and $x_f = G(V_{LR})$ denotes generated SR volumes. The discriminator loss ($L_D$) is correspondingly:

\begin{equation}
\begin{split}   
L_D = -\mathbb{E}_{x_r}[\log(\sigma(C(x_r) - \mathbb{E}_{x_f}[C(x_f)]))] \\ - \mathbb{E}_{x_f}[\log(1 - \sigma(C(x_f) - \mathbb{E}_{x_r}[C(x_r)]))]
\end{split}
\label{eq:discriminator_loss}
\end{equation}

Training involves optimizing $G$ and $D$ iteratively using these loss functions, typically with patch-based training on 3D sub-volumes due to computational constraints.

\subsection{Evaluation Metrics}
To evaluate the performance of super-resolution (SR) models, we employ several quantitative metrics that assess the fidelity of reconstructed high-resolution (HR) images relative to ground truth (GT) images. Although visual inspection provides some insight into image quality, quantitative metrics offer more precise and reproducible evaluations. In this work, we used multiple metrics, including Peak Signal-to-Noise Ratio (PSNR), Multi-Scale Structural Similarity Index Measure (MS-SSIM), Visual Information Fidelity (VIF), and Mean Squared Error (MSE) during the model training phase. Each metric measures distinct aspects of image quality, enabling a more comprehensive assessment of the model’s ability to recover fine details and preserve perceptual realism.
\subsubsection{Visual Information Fidelity (VIF)}
unlike PSNR and MS-SSIM which used in multiple publications for SR \cite{p_esrgan,p_medvae}, VIF is not used before and we used it in this paper to be able to have this metric adaptable to medical data. It is another image quality metric that models the human visual system better than PSNR or SSIM. It is based on the idea that image quality can be measured by the amount of mutual information between the reference (GT) image and the distorted (HR) image. VIM assumes that image distortions introduce a loss of information that affects the perceived quality.
\begin{equation}
    \text{VIF} = \frac{\sum_{j \in \text{sub}} I(C_j; F_j | s_j)}{\sum_{j \in \text{sub}} I(C_j; E_j | s_j)}
    \label{eq:vif}
\end{equation}
where $C_j$ represents the coefficients of the reference image, $F_j$ represents the coefficients of the distorted image, and $E_j$ represents the coefficients of the reference image in $j-th$ subband after passing through a hypothetical ideal channel. All coefficients are then summed up over different subbands.
It measures the amount of visual information that can be extracted from the distorted image relative to the reference image by an ideal observer. In essence, it measures how much of the original information survives the distortion process.
A higher VIF score indicates that the reconstructed high-resolution (HR) image retains more of the original visual information, suggesting better perceptual fidelity. VIF is particularly effective in capturing degradations such as blur, noise, and compression artifacts, which often impair image quality in ways that closely align with human perception. Given the prevalence of noise and blurring in brain imaging data, we implemented and employed VIF to evaluate our model’s ability to preserve perceptually relevant information.

\section{Experiments}
\subsection{Dataset}
This study utilized data from the Adolescent Brain Cognitive Development (ABCD) Study, the largest long-term investigation of brain development and child health in the United States. The findings presented are based on data from release 5.1 of the ABCD dataset. This dataset collected over 11,800 children, initially aged 9-10 years at baseline. The ABCD study's data collection covers a broad spectrum of measures, including neurocognitive batteries, physical and mental health assessments, and other relevant health backgrounds, all of which are used in finding relationships between adolescent behavior and brain function\cite{zening1,zening2,p_fmri_data,p_cnn_mdl}.
Access to the ABCD data for this study was granted under Application ID 13591. Fast Track imaging data were downloaded from the National Data Archive (NDA). All data collection procedures adhered to rigorous ethical standards, including written informed consent from parents and assent from children. These protocols were approved by a centralized Institutional Review Board (IRB) at the University of California, San Diego, with local IRB approval obtained at each participating site\cite{zening1,zening2}.

\subsection{preprocessing}
In this dataset, resting-state fMRI data from the ABCD dataset applied pre-processing pipeline, applied a combination of the FMRIB Software Library (FSL) v6.0 toolbox and the Statistical Parametric Mapping (SPM) 12 toolbox within the MATLAB 2020b environment. This pre-processing sequence involved several key steps including rigid body motion correction, distortion correction, removal of dummy scans, normalization to the standard Montreal Neurological Institute (MNI) space, and spatial smoothing with a 6 mm Gaussian kernel. Following pre-processing, data quality control on the fMRI data was meticulously performed. For the within-session analysis, a total of 9,071 participants from the baseline session and 2,918 participants from the second-year session met the criteria. In the context of cross-session analysis, 2,290 participants were identified with four good scans from both the baseline and second-year sessions.
Using the ABCD dataset, we extracted resting-state functional connectivity (FC) spatial components to serve as ground truth (GT) using the NeuroMark framework, a spatially constrained independent component analysis (ICA) method. Each component volume has dimensions of $X=53$, $Y=63$, and $Z=52$. To generate low-resolution (LR) data, we downsampled the GT components using bilinear interpolation with a scaling factor of 0.5, thereby reducing the spatial resolution of the data. The primary objective of this study is to train an optimized generator model that upscales these LR inputs by a factor of 2 to produce high-resolution (HR) reconstructions that closely resemble the original GT components.
To improve the generalizability of our model to real-world medical fMRI data, we further augmented the LR inputs by adding Rician noise , type of noise in MRI acquisitions. Rician noise arises from the magnitude of complex-valued MR signals. It applied as follows:
\begin{equation}
    P(x|\nu, \sigma) = \frac{x}{\sigma^2} \exp\left(-\frac{x^2 + \nu^2}{2\sigma^2}\right) I_0\left(\frac{x\nu}{\sigma^2}\right)
\label{eq_rician}
\end{equation}
where $x$ is the observed magnitude value, $\nu$ is the true signal amplitude, $\sigma$ is the standard deviation of Gaussian noise, and $I_0$ is Bessel function of the first kind with order zero. applying this noise simulates more realistic neuroimaging conditions and it aims to robust model learning.
 
\textbf{Data Splitting:} The entire dataset of spatial maps, derived from individual subjects, was partitioned to ensure subject-level separation between sets, thereby preventing data leakage. We allocated 90\% of the subjects' data to the training set, 5\% to the validation set, and the remaining 5\% to the test set. The training set was used to fit the model parameters. The validation set was crucial for hyperparameter tuning, including optimizing the weighting coefficients ($\lambda_{pixel}$, $\lambda_{perc}$, $\lambda_{adv}$) for the composite loss function and determining the optimal number of training epochs. The test set was held out entirely and used only for the final, unbiased evaluation of the trained model's performance.

\textbf{Hyperparameters and Architecture:} The NeuroGAN-3D generator was constructed with 16 RRDB-3D blocks. Due to the high memory requirements of processing full 3D volumes, training was performed on randomly extracted 3D patches of size $32 \times 32 \times 32$. We used the Adam optimizer with $\beta_1 = 0.9$ and $\beta_2 = 0.999$.

\textbf{Training Procedure:} The training process was implemented in two distinct stages to ensure stability and enhance detail generation. 
\begin{enumerate}
    \item \textbf{Generator Pre-training:} The generator was first trained for 200 epochs using only the pixel-wise L1 loss. This stage focused on restoring the primary content of the images. The learning rate was initialized to $2 \times 10^{-4}$.
    \item \textbf{Adversarial Training:} Subsequently, the pre-trained generator and the discriminator were trained jointly for 130 epochs using the full composite loss function. The learning rate was again initialized to $2 \times 10^{-4}$ and was halved every 40 epochs to facilitate fine-tuning and stable convergence.
\end{enumerate}
The models were trained on a high-performance computing cluster equipped with an NVIDIA A100 GPU, 16 CPU cores, and 80GB of RAM.

\begin{figure*}[ht]
  \includegraphics[width=\textwidth]{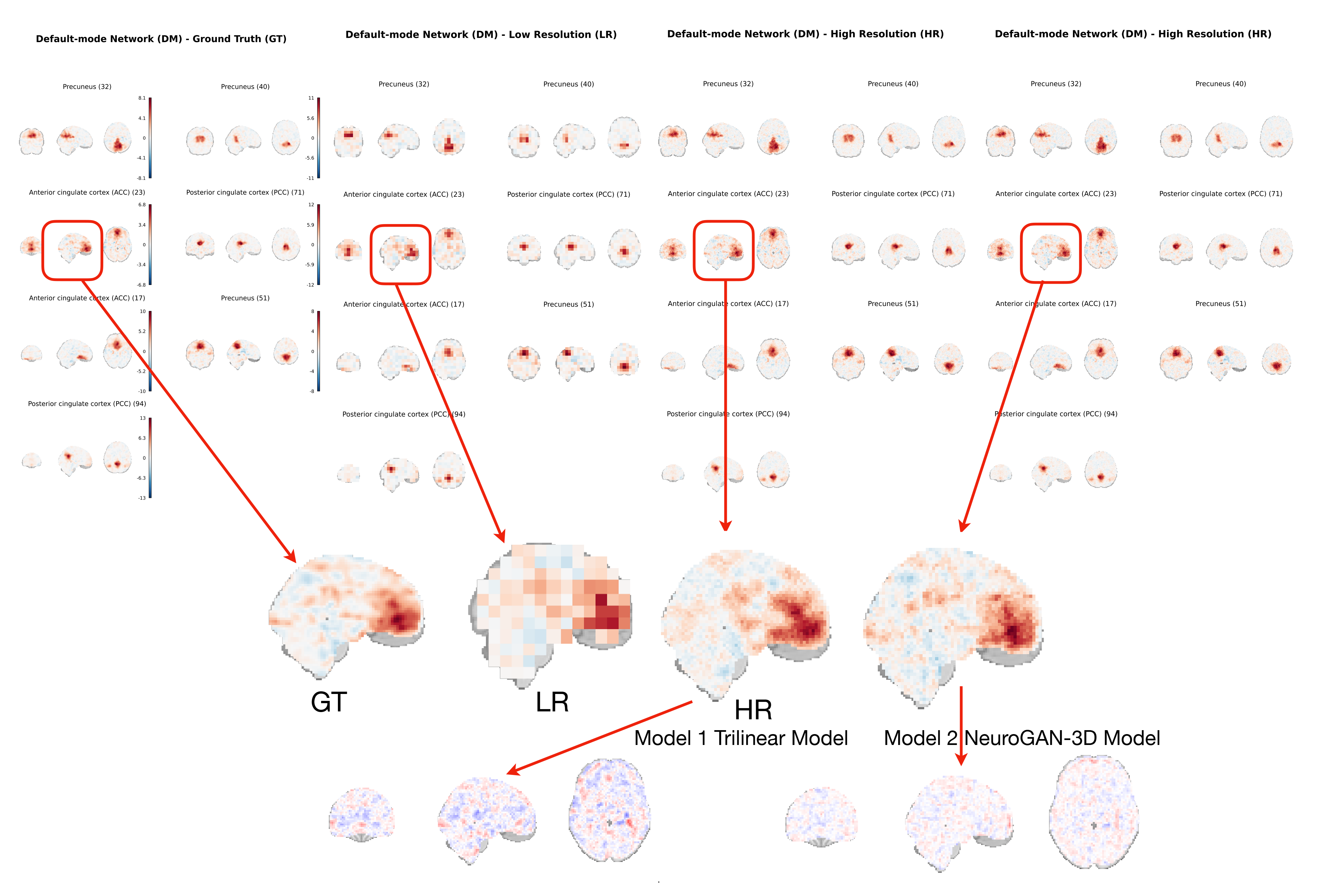}
  \caption{Visualization of DMN components results. We show the GT data. The second column shows LR images as input to the models. The third and fourth columns present the HR outputs from Model 1 (trilinear model) and Model 2 (NeuroGAN-3D), respectively. Bars are same for all images. Magnified views of selected regions, indicated by red boxes, are provided at the bottom to highlight the differences in detail and then showing the diffrence between the LR input, GT, and HR outputs from both models.}
\label{fig:visual_comparison}
\end{figure*} 

\section{Results}
In this part, we want to evaluate our proposed model, NeuroGAN-3D, against a baseline model(in this study as trilinear model based on\cite{p_medvae}) and compare its performance metrics with GT components. the aim is showing the efficacy of NeuroGAN-3D in enhancing the resolution of 3D data.
The performance of our proposed model (referred to as Model 2: NeuroGAN-3D) and a baseline model (Model 1: Trilinear Model) was measured using PSNR, MS-SSIM, VIF, and MSE. The evaluation focused on key components of the Default Mode Network (DMN): the Precuneus, Anterior Cingulate Cortex (ACC), and Posterior Cingulate Cortex (PCC). While we focus on the DMN for detailed visualization in this paper, it is important to note that the quantitative metrics reported in Table \ref{tab:results} were calculated by averaging performance across all 53 identified resting-state networks, confirming the model's generalizability beyond the DMN. Training and evaluation process applied to other networks along with DMN.
As detailed in Table \ref{tab:results}, Model 2 outperformed Model 1 in metrics in all DMN components. Specifically, Model 2 achieved overall average PSNR of 31.8 dB, compared to 29.6 dB for Model 1. Similarly, it obtained a higher average SSIM (0.94 vs. 0.89), higher average VIF (0.76 vs. 0.66), and a lower average MSE (0.001 vs. 0.002). The higher scores for PSNR, SSIM, and VIF indicate a better reconstruction quality and perceptual fidelity, while the lower MSE signifies a smaller error between HR and GT volumes.
Figure \ref{fig:visual_comparison} provides a detailed visual comparison of SR outputs for the DMN components. The top row of Figure \ref{fig:visual_comparison} displays various brain slices, showcasing the original GT data, the LR input, and HR outputs generated by both Model 2 (NeuroGAN-3D) and Model 1 (Trilinear Model). The LR images appear visibly pixelated and lack the fine spatial details present in the GT images, particularly evident in the blurred boundaries of functional activation. In contrast, the HR output from Model 2 (NeuroGAN-3D) demonstrates improvement in image quality, closely resembling the GT. also as an instance, observing the magnified views in the bottom row of Figure \ref{fig:visual_comparison}, the specific DMN region highlighted by the red box reveals that NeuroGAN-3D (Model 2) successfully restores intricate structural details and produces sharp, coherent spatial maps, closely mirroring the GT. The reconstructed activation patterns exhibit clear boundaries and a faithful representation of the underlying anatomy. In contrast, the output from Model 1 (Trilinear Model), while an improvement over the LR input, shows visibly less detail, introduces blurring, and exhibits a lack of realism compared to Model 2, especially in the fine-grained activation patterns.

\subsection{Ablation Study}
To validate the contributions of individual components within our proposed NeuroGAN-3D model, we conducted an ablation study. We compared the complete NeuroGAN-3D model (Model 2) against an ablated baseline model (Model 1). Model 2 represents NeuroGAN-3D architecture, which includes the RRDB-3D generator, the PatchGAN discriminator, and a composite loss function featuring a novel SSIM-based perceptual loss. Model 1 is an ablated version of the framework, relying solely on the pixel value wise losses. This comparison allows us to isolate and quantify the impact of our model.
The results of this comparison are summarized in Table \ref{tab:results}. It demonstrates that Model 2 outperforms Model 1 across evaluation metrics. In the overall average across DMN components, Model 2 achieves a PSNR of 31.8 dB and an SSIM of 0.94, which are substantially higher than Model 1 with PSNR of 29.6 dB and SSIM of 0.89.
we should note that, the largest improvements are observed in the perceptually-oriented metrics. Model 2 with VIF score of 0.76 is higher than Model 1 with score of 0.66. Since VIF and SSIM are designed to measure structural and perceptual similarity that aligns with human visual perception, this performance gap confirms the role of the $L_{perc}$ loss term. By explicitly optimizing for structural fidelity via the novel SSIM function, the generator is also guided to produce results that are more visually and structurally plausible.
The results presented in Figure \ref{fig:visual_comparison} further support this conclusion. Specifically, when comparing Model 2 NeuroGAN-3D and Model 1 Trilinear Model outputs in the lower magnified view of Figure \ref{fig:visual_comparison}, the HR output from Model 2 is visually sharper, exhibits clearer boundaries of functional activation, and displays more intricate details than the output from Model 1. the reconstruction appears comparatively less defined, with blurring and a lack of structural fidelity present in Model 1.

\section{Discussion}
In this paper, we introduced NeuroGAN-3D, a novel framework for high-fidelity super-resolution of 3D neuroimaging data. Inspired by ESRGAN, our model adapts and extends its principles for volumetric 3D data, addressing the inherent limitations of conventional slice-by-slice 2D methods, which often lead to a loss of inter-slice continuity and the introduction of undesirable artifacts. Our results demonstrate the potential of NeuroGAN-3D in improving the resolution of rs-fMRI spatial maps, significantly outperforming the proposed baseline model.

The strong performance of NeuroGAN-3D can be attributed to a combination of its 3D architecture, the composite loss function, and a carefully designed model development pipeline. Our data splitting strategy, which strictly separated subjects into training, validation, and testing sets, was fundamental for obtaining an unbiased evaluation of the model's generalization capabilities. Furthermore, we incorporated a data augmentation step by adding Rician noise to the low-resolution inputs. This was a critical choice designed to simulate the noise characteristics inherent in real-world MRI acquisitions, thereby improving the model's robustness and its applicability to clinical data.
The training process itself was implemented in two strategic stages to ensure stable convergence and high-fidelity results. The initial pre-training of the generator for 200 epochs using a simple pixel-wise L1 loss allowed the model to first learn the basic structure and content of the brain maps. This was followed by a 130-epoch adversarial training phase, where the full composite loss function guided the model to refine textures and generate perceptually realistic high-frequency details. This two-stage approach is crucial for mitigating the instability often associated with training GANs from scratch and was instrumental in achieving the final quality of our reconstructions.
As visually shown in Figure \ref{fig:visual_comparison}, NeuroGAN-3D (Model 2) generates results that more like to the GT in terms of structural integrity and detailed functional activation patterns compared to model1. This visual fidelity directly correlates with the higher perceptual metrics (SSIM, VIF) achieved by model.

Despite the promising results, this study has several limitations. A primary limitation, as noted, is that the model was trained and validated exclusively on rs-fMRI data from a single, albeit large, dataset (ABCD). Consequently, the model's performance on data acquired from different scanners, using different acquisition parameters, or from different populations remains unevaluated. This lack of external validation means the generalizability of NeuroGAN-3D is not yet fully established. Future work must prioritize testing the model on external, publicly available neuroimaging datasets to rigorously assess its robustness and broader applicability. Furthermore, its current application is limited to functional spatial maps; extending and validating its performance on other modalities, such as structural MRI or diffusion tensor imaging, is another critical next step. Additionally, the loss weighting coefficients were determined empirically. While our chosen weights are justified by the model's strong performance, we acknowledge that this study lacks a formal ablation study or a systematic hyperparameter search (e.g., grid search) to explore the sensitivity of the model to these parameters. A more rigorous optimization of these weights could potentially yield further improvements and is an important direction for future work.

\begin{table}[ht]
    \centering
    \caption{Performance of SR Models Across DMN Components. The table details performance on a representative set of individual DMN spatial map components from the test set. The "Overall Average" and "Standard Deviation (STD)" summarize the performance across all test components from an optimized training run. Model stability was confirmed by repeating the training process with multiple random seeds, which yielded consistent performance within a narrow margin of the reported averages.}
    \label{tab:results}
    \newcolumntype{L}{>{\raggedright\arraybackslash}p{2.8cm}}
    \small
    \resizebox{\columnwidth}{!}{%
\begin{tabular}{l l c c c c}
    \toprule
    \textbf{Model} & \textbf{DMN Component} & \textbf{PSNR} ($\uparrow$) & \textbf{SSIM} ($\uparrow$) & \textbf{VIF} ($\uparrow$) & \textbf{MSE} ($\downarrow$) \\
    \midrule
    \multirow{9}{*}{Model 2}
    & Precuneus & 32.4 & 0.95 & 0.78 & 0.001 \\
    & Precuneus & 31.9 & 0.94 & 0.76 & 0.001 \\
    & ACC       & 30.7 & 0.93 & 0.74 & 0.002 \\
    & PCC       & 32.1 & 0.95 & 0.77 & 0.001 \\
    & ACC       & 31.5 & 0.94 & 0.75 & 0.001 \\
    & Precuneus & 31.8 & 0.94 & 0.76 & 0.001 \\
    & PCC       & 32.2 & 0.95 & 0.78 & 0.001 \\
    \cmidrule{2-6}
    & \textbf{Overall Average} & \textbf{31.8} & \textbf{0.94} & \textbf{0.76} & \textbf{0.001} \\
    & \textbf{STD} & $\pm$0.52 & $\pm$0.007 & $\pm$0.014 & $\pm$0.00035 \\
    \midrule
    \multirow{9}{*}{Model 1}
    & Precuneus & 30.1 & 0.90 & 0.68 & 0.002 \\
    & Precuneus & 29.7 & 0.89 & 0.66 & 0.002 \\
    & ACC       & 28.5 & 0.88 & 0.64 & 0.003 \\
    & PCC       & 29.9 & 0.90 & 0.67 & 0.002 \\
    & ACC       & 29.3 & 0.89 & 0.65 & 0.002 \\
    & Precuneus & 29.6 & 0.89 & 0.66 & 0.002 \\
    & PCC       & 30.0 & 0.90 & 0.68 & 0.002 \\
    \cmidrule{2-6}
    & \textbf{Overall Average} & \textbf{29.6} & \textbf{0.89} & \textbf{0.66} & \textbf{0.002} \\
    & \textbf{STD} & $\pm$0.51 & $\pm$0.007 & $\pm$0.014 & $\pm$0.00035 \\
    \bottomrule
\end{tabular}%
}
\end{table}

\section{Discussion}
In this paper, we introduced NeuroGAN-3D, a novel framework for high-fidelity super-resolution of 3D neuroimaging data. Inspired by ESRGAN, our model adapts and extends its principles for volumetric 3D data, addressing the inherent limitations of conventional slice-by-slice 2D methods, which often lead to a loss of inter-slice continuity and the introduction of undesirable artifacts. Our results demonstrate the potential of NeuroGAN-3D in improving the resolution of rs-fMRI spatial maps, significantly outperforming the proposed baseline model.

The strong performance of NeuroGAN-3D can be attributed to a combination of its 3D architecture, the composite loss function, and a carefully designed model development pipeline. Our data splitting strategy, which strictly separated subjects into training, validation, and testing sets, was fundamental for obtaining an unbiased evaluation of the model's generalization capabilities. Furthermore, we incorporated a data augmentation step by adding Rician noise to the low-resolution inputs. This was a critical choice designed to simulate the noise characteristics inherent in real-world MRI acquisitions, thereby improving the model's robustness and its applicability to clinical data.
The training process itself was implemented in two strategic stages to ensure stable convergence and high-fidelity results. The initial pre-training of the generator for 200 epochs using a simple pixel-wise L1 loss allowed the model to first learn the basic structure and content of the brain maps. This was followed by a 130-epoch adversarial training phase, where the full composite loss function guided the model to refine textures and generate perceptually realistic high-frequency details. This two-stage approach is crucial for mitigating the instability often associated with training GANs from scratch and was instrumental in achieving the final quality of our reconstructions.
As visually shown in Figure \ref{fig:visual_comparison}, NeuroGAN-3D (Model 2) generates results that more like to the GT in terms of structural integrity and detailed functional activation patterns compared to model1. This visual fidelity directly correlates with the higher perceptual metrics (SSIM, VIF) achieved by model.
Despite the promising results, this study has several limitations. A primary limitation, as noted, is that the model was trained and validated exclusively on rs-fMRI data from a single, albeit large, dataset (ABCD). Consequently, the model's performance on data acquired from different scanners, using different acquisition parameters, or from different populations remains unevaluated. This lack of external validation means the generalizability of NeuroGAN-3D is not yet fully established. Future work must prioritize testing the model on external, publicly available neuroimaging datasets to rigorously assess its robustness and broader applicability. Furthermore, its current application is limited to functional spatial maps; extending and validating its performance on other modalities, such as structural MRI or diffusion tensor imaging, is another critical next step. Additionally, the loss weighting coefficients were determined empirically. While our chosen weights are justified by the model's strong performance, we acknowledge that this study lacks a formal ablation study or a systematic hyperparameter search (e.g., grid search) to explore the sensitivity of the model to these parameters. A more rigorous optimization of these weights could potentially yield further improvements and is an important direction for future work.

\section{Conclusion}
In this paper, we introduced NeuroGAN-3D, modified generative adversarial network architecture specifically designed to improve volumetric 3D neuroimaging data with higher fidelity. Recognizing the limitations of  2D-based methods and the constraints of direct HR data, our model aims to implement a 3D framework to ensure inter-slice continuity and preserve the structural details in brain data. Our loss function is optimized to produce outputs that are not only accurate but also anatomically and functionally trustworthy.
The contributions of this work are threefold: the development of a novel 3D GAN architecture tailored for neuroimaging data, the implementation of a modified loss function to ensure high-fidelity reconstructions, and the successful application of this framework to enhance spatial maps critical for connectomic analysis. Future work will focus on extending the model to other imaging modalities. Finally, based on the results, NeuroGAN-3D provides a robust and effective tool to push the boundaries of neuroimaging research.

\begin{credits}
\subsubsection{\ackname} This work was supported in part by the National Science Foundation (NSF) under Grant 2112455 and in part by the National Institutes of Health (NIH) through Grants R01MH123610 and R01MH136665.

\subsubsection{\discintname}
  The authors have no other competing interests to declare that are relevant to the content of this article.
\end{credits}

\end{document}